\documentclass[letterpaper]{article} %
\usepackage[draft]{aaai2026}  %
\usepackage{times}  %
\usepackage{helvet}  %
\usepackage{courier}  %
\usepackage[hyphens]{url}  %
\usepackage{graphicx} %
\usepackage{natbib}  %
\usepackage{caption} %
\usepackage{algorithm}
\usepackage{algorithmic}

\usepackage{newfloat}
\usepackage{listings}
\DeclareCaptionStyle{ruled}{labelfont=normalfont,labelsep=colon,strut=off} %
\floatstyle{ruled}
\newfloat{listing}{tb}{lst}{}
\floatname{listing}{Listing}
\usepackage{annotate-equations}
\usepackage{subcaption}
\usepackage{pifont}%
\usepackage{placeins}
\usepackage{tabularx}
\usepackage{booktabs} %
\usepackage{soul}
\usepackage{multirow}
\usepackage{adjustbox}
\usepackage{xspace} 
\usepackage{amsmath} 
\usepackage{colortbl}
\usepackage{amsthm}
\usepackage[inkscapelatex=false]{svg}

\makeatletter
\DeclareRobustCommand\onedot{\futurelet\@let@token\@onedot}
\def\@onedot{\ifx\@let@token.\else.\null\fi\xspace}
\def\eg{\emph{e.g}\onedot} 
\def\ie{\emph{i.e}\onedot}

\makeatother

\def\ourSDS{AnchorDS} 

\definecolor{myorange}{HTML}{B13B12}
\definecolor{myblue}{HTML}{14A0EC}
\definecolor{mygreen}{HTML}{7ED569}
\definecolor{mygred}{HTML}{F8D5D5}
\usepackage[pagebackref,breaklinks,colorlinks,citecolor=myblue]{hyperref}

\def\eg{\emph{e.g}}

\title{AnchorDS: Anchoring Dynamic Sources for\\ Semantically Consistent Text-to-3D Generation}

\author {
    Jiayin Zhu\textsuperscript{\rm 1},
    Linlin Yang\textsuperscript{\rm 2},
    Yicong Li\textsuperscript{\rm 1}\thanks{Corresponding author.},
    Angela Yao\textsuperscript{\rm 1},
}
\affiliations {
    \textsuperscript{\rm 1}National University of Singapore\\
    \textsuperscript{\rm 2}Communication University of China\\
    zhujiayin@u.nus.edu, lyang@cuc.edu.cn, liyicong@u.nus.edu, ayao@comp.nus.edu.sg
}

\begin{document}

\maketitle

\begin{abstract}
Optimization‐based text‑to‑3D methods distill guidance from 2D generative models via Score Distillation Sampling (SDS), but implicitly treat this guidance as static.   %
This work shows that ignoring source dynamics yields inconsistent trajectories that suppress or merge semantic cues, leading to ``semantic over-smoothing'' artifacts.  As such, we reformulate text‑to‑3D optimization as mapping a \emph{dynamically evolving} \emph{source} distribution to a fixed target distribution.  We cast the problem into a dual‑conditioned latent space, conditioned on both the text prompt and the intermediately rendered image.  {Given this joint setup,} we observe that the image condition naturally anchors the current source distribution.  
Building on this insight, we introduce \ourSDS, an improved score distillation mechanism %
that provides {state‑anchored guidance} with image conditions and stabilizes generation.
We further %
penalize erroneous source estimates and design a lightweight filter strategy and %
fine‑tuning strategy that refines the anchor with negligible overhead.  
\ourSDS\ produces finer-grained detail, more natural colours, and stronger semantic consistency, particularly for complex prompts, while maintaining efficiency. Extensive experiments show that our method surpasses previous methods in both quality and efficiency. %
\end{abstract}

\begin{links}
    \link{Code}{https://github.com/viridityzhu/AnchorDS}
\end{links}

\section{Introduction}
\label{sec:intro}

With the growing demand for 3D content creation in gaming and virtual reality, text-to-3D generation has emerged as a significant research frontier.  One prominent solution that builds on this trend is Score Distillation Sampling (SDS)~\cite{poole_dreamfusion_2022}.  SDS can leverage powerful off-the-shelf 2D diffusion models to guide optimization-based text-to-3D generation without large-scale 3D datasets. 

\begin{figure}[t]
    \centering
    \includegraphics[width=\linewidth]{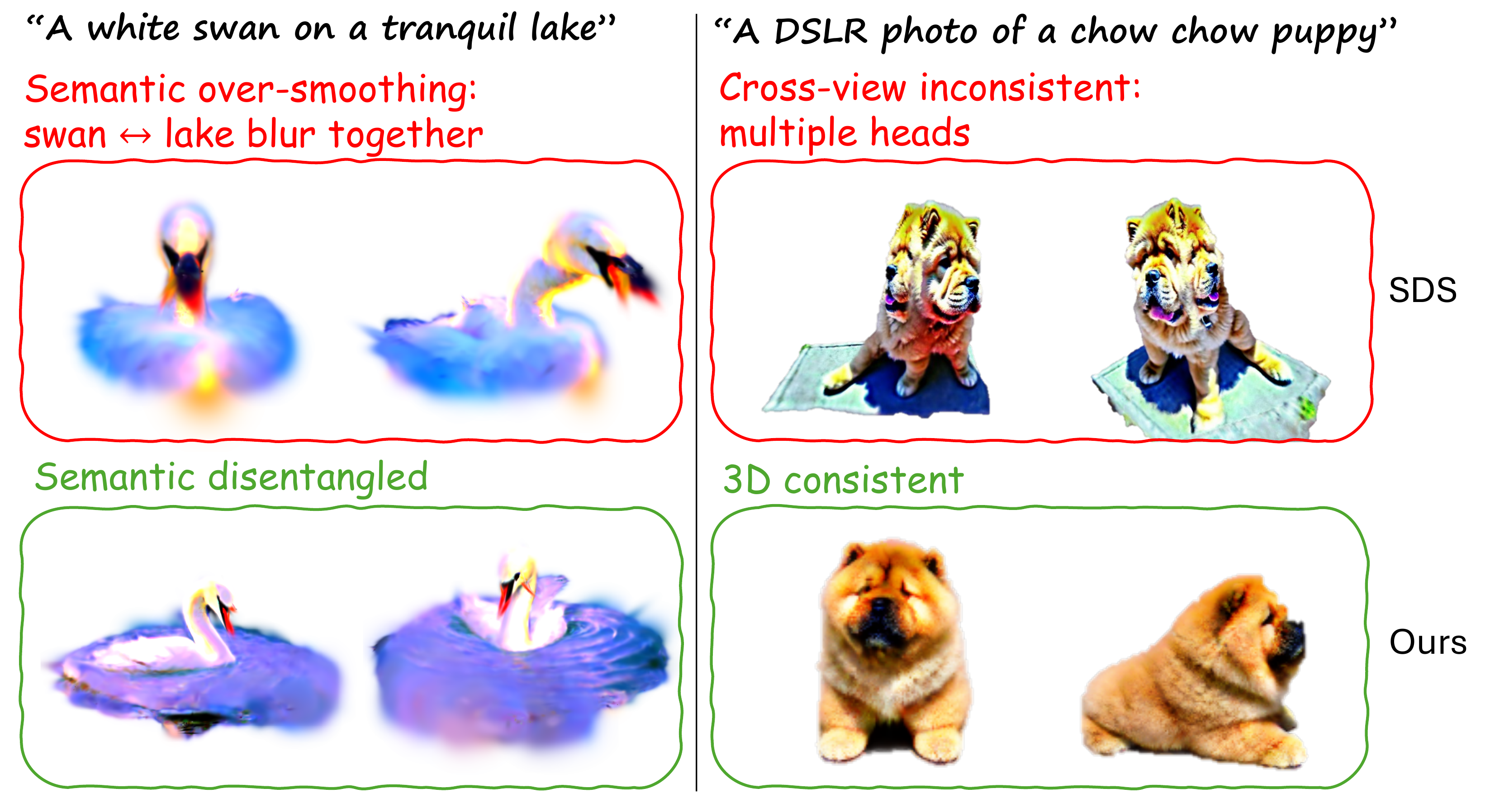}
    \vspace{-22pt}
    \caption{\small\textbf{Comparison of Vanilla SDS vs. Ours.} SDS suffers from semantic over-smoothing (mixing swan and lake) and cross-view inconsistency (multiple heads). Ours achieves semantic disentanglement and 3D consistency across views.}
    \label{fig:teaser}
    \vspace{-10pt}
\end{figure}

SDS-guided text-to-3D generation can be interpreted as an optimization process that gradually shifts the distribution of rendered images from the current 3D representation (\ie, the source distribution) toward a distribution defined by a text-conditioned diffusion model (\ie, the target distribution)~\cite{mcallister2024sdsbridge}. Under this interpretation, {follow-up works have proposed enhancements} follow two strategies. The first enhances the target distribution estimation by incorporating additional conditions, such as multiview images ~\cite{shi2023mvdream,li2025mvcontrol} or depth maps~\cite{qiu2024richdreamer}. 
The second addresses source distribution ismatch by refining the guidance, using reference-based scoring~\cite{Hertz_2023_dds}, enhanced negative prompts~\cite{mcallister2024sdsbridge}, or variational optimization~\cite{wang_prolificdreamer_2023}. %
{Despite these advancements, SDS-based 3D generation still struggles with consistently preserving structural semantics across iterative updates. This manifests as two critical artifacts: (1) semantic over-smoothing, where object-specific features degenerate into homogenized, semantically ambiguous representations; and (2) cross-view inconsistency, where geometry and appearance are incoherent across perspectives} (Fig.~\ref{fig:teaser}).

At its core, SDS optimization is formulated as a %
distribution transformation: it progressively shifts the rendered appearance of the 3D asset—from a \textit{source distribution} reflecting the current 3D state toward a \textit{target distribution} defined by a text-conditioned diffusion model. 
However, while the target distribution remains fixed to the text prompt, the source distribution is {approximated} using a static unconditional prior throughout optimization. {This critically overlooks the \textit{inherently non-stationary nature of the source distribution}: as the 3D representation evolves through optimization steps, rendered images continuously update, dynamically altering the source distribution. Consequently, SDS effectively discards accumulated structural information at each step by restarting from a semantics-agnostic prior. This fundamental mismatch between the static guidance mechanism and the evolving asset state destabilizes optimization, ultimately manifesting as the observed artifacts.} %

To address this issue, we propose to reframe score distillation as a dynamic editing process. 
Rather than treating generation as a one-shot projection from a static source, we view it as a progressive editing loop in which each optimization step refines the current 3D state based on the accumulated guidance from previous steps. 
Crucially, by explicitly recognizing and exploiting the evolving 3D state itself, our reformulation yields two advantages:
1) it supplies a stream of rich cues—geometry, colour, and semantics—that stabilise guidance and enforce cross-view consistency;
and 2) feeding the state back into a pretrained conditional diffusion model enables dynamic, accurate, and lightweight source estimates without extra networks or handcrafted prompts.

Building on this perspective, we introduce \ourSDS, a dynamic form of SDS %
that anchors the source distribution using the rendered image at each optimization step. Specifically, we leverage a dual-conditioned diffusion model that incorporates both the text prompt and the intermediate image as guidance signals. Crucially, we observe that the diffusion model's predicted noise conditioned on the current rendering naturally encodes structural and semantic cues from the image condition. This noise prediction inherently anchors the source distribution by correlating the gradient guidance with the evolving 3D state, mitigating distribution drift without explicit constraints. Notably, the image condition does not constrain the target output directly, but serves as a contextual anchor that steers the generation. This design is not sensitive to the selection of the dual-conditioned model.  Our method is robust when utilizing various popular image-conditioned adapters, \eg, IP-Adapter~\cite{ye2023ipadapter} and ControlNet~\cite{zhang2023controlnet}. %

{Despite this flexibility, anchoring dynamic sources is still non-trivial because the implicit latent space of the diffusion model exhibits a distribution mismatch with rendered images. To bridge the gap,} 
we incorporate two complementary components. First, we reconstruct a pseudo-source image at each step, offering a metric for evaluating the quality of the estimated source distribution. Second, we introduce two practical enhancements: a simple yet effective \textit{Filtering} mechanism that discards unreliable source predictions, and a lightweight \textit{Fine-tuning} strategy that better aligns the diffusion model with the domain of rendered images.

Our contributions are summarized as follows:
\begin{enumerate}
    \item We reveal the evolving nature of the source distribution in SDS and identify it as the root cause of semantic oversmoothing and inconsistent optimization trajectories.

    \item We propose \ourSDS, a novel score distillation framework that dynamically anchors the source estimation via a dual-conditioned diffusion model. We further introduce a filtering mechanism and a lightweight fine-tuning strategy to better regularize the evolving source distribution.

    \item Extensive experiments on T$^3$Bench~\cite{he2023t3bench} and {a representative suite of challenging prompts} demonstrate that our method outperforms state-of-the-art (SoTA) SDS variants in both generation quality and efficiency.
\end{enumerate}

\section{Related Works}
\noindent\textbf{Text-Guided 3D Generation.}
DreamFusion demonstrates that a NeRF can be trained from scratch by following Score Distillation Sampling (SDS) gradients derived from 2D diffusions~\cite{poole_dreamfusion_2022}.  
Subsequent works keep the same optimization-from-noise paradigm while improving efficiency or fidelity~\cite{lin2023magic3d,tang_dreamgaussian_2023,yi2023gaussiandreamer,li_sweetdreamer_2023}. 
Although these methods deliver the highest visual quality, they remain slow and are still vulnerable to view inconsistency and Janus problems.  
To eliminate per-scene optimization, another line of work trains generators that map text directly to 3D latents~\cite{jun2023shap_e,Nichol2022PointE,siddiqui2024meta3dgen,xiang2024trellis}.
While generation is fast, training quality rely on large-scale 3D datasets~\cite{objaverse,Fu_2021_3d_future,Collins_2022_abo}, and the outputs still trail optimization methods in geometric and color accuracy%
~\cite{he2023t3bench}.  
Hybrid strategies therefore combine a feed-forward initialization with subsequent Gaussian/NeRF refinement~\cite{liang2024luciddreamer,GaussianDreamerPro, yi2023gaussiandreamer}.  

\noindent\textbf{Conditional and Controllable Text-to-3D.}
Extending the diffusion prior with extra modalities improves controllability. ControlNet~\cite{zhang2023controlnet} has been adopted for depth, normal, or multi-view constraints~\cite{huangcvpr24dreamcontrol,li2025mvcontrol}. IP-Adapter \cite{ye2023ipadapter} let users steer style or identity with a reference image and have been plugged into 3D pipelines~\cite{Zeng2023IPDreamer}. All these works, however, treat the additional image as an external positive condition supplied a priori. We instead employs the intermediate rendering itself as a dynamic \emph{source} anchor supplying self-consistent guidance.%

\noindent\textbf{Refining Score Distillation Sampling.}
Recent work revisits SDS from theoretical and practical perspectives. 
\cite{yu_csd_2023,tang2023stable,katzir2024nfsd} separate mode-seeking and variance terms to stabilize optimization.  
\cite{liang2024luciddreamer,lukoianov2024sdi} avoid first-order errors through DDIM sampling and inversion.  
Mismatch remedies attempt to align the diffusion prior with the 3D asset. DDS pairs each original image with a reference prompt~\cite{Hertz_2023_dds}, SDS-Bridge introduces a handcrafted prompt describing the poor 3D state~\cite{mcallister2024sdsbridge}, and ProlificDreamer trains a LoRA branch to approximate the particle distribution~\cite{wang_prolificdreamer_2023}.  
These solutions improve stability yet rely on static prompts, specific references, or auxiliary networks, which introduce new bias and overhead.  
Instead, we remove this dependency by directly feeding the current rendering into the diffusion prior, yielding faithful and bias-free guidance efficiently.%

\section{Analysis on the Issue of Source Distribution Estimation}\label{sec:source_estimation_issue}
We begin by revisiting the formulation of text-to-3D generation via Score Distillation Sampling (SDS)~\cite{poole_dreamfusion_2022}, and highlight its key limitation - a lack of awareness of the rendered appearance from the current 3D state. We then draw connections between SDS and 2D editing paradigms, and reinterpret SDS as a dynamic editing process that conditions on the evolving source.

\subsection{Preliminaries}\label{sec:revisit_3d_gen}

\noindent\textbf{Score Distillation Sampling (SDS).} 
SDS leverages pre-trained 2D diffusion models to optimize the 3D generation. Specifically, it applies the denoising process of a 2D diffusion model to a rendered image from the 3D model, through which it distills a %
prior on the generated 3D output.  
Given a sampled noise $\epsilon\sim \mathcal{N}(0,\mathbf{I})$ and a latent representation $z$ of the rendered image, the noisy latent at timestep $t$ is given by 
\begin{equation}\label{eq:z_0_to_z_t}
    z_t = \sqrt{\bar\alpha_t}\,z + \sqrt{1-\bar\alpha_t}\,\epsilon.
\end{equation}
The corresponding noisy image is then used to compute the SDS gradient $\nabla_\Theta \mathcal{L}_{SDS}(\phi, z)$, where the %
predicted noise is assumed to approximate the score function. Concretely, {following the score matching view, the diffusion model $\phi$'s noise prediction $\hat{\boldsymbol \epsilon}_{\phi}$ is defined as}:
\begin{equation}\label{eq:p_phi}
    \hat{\boldsymbol\epsilon}_\phi = -\sigma_t \nabla_{z_t}\log p(z_t;t,c), \quad \text{with } \sigma_t=\sqrt{1-\bar\alpha_t},
\end{equation}
where $c$ denotes conditioning information (typically a text prompt $y$, or a combination $c=\{y, I\}$ when an image condition $I$ is included). This formulation directs the optimization toward regions of high density in the conditional distribution $p(z_t;t,c)$.
Applied to a 3D model parameterized by $\Theta$, the gradient is given as:
\begin{equation}\label{eq:van_sds}
     \nabla_\Theta \mathcal{L}_{SDS}(\phi, z)=w(t)\left(\hat{\boldsymbol{\epsilon}}_\phi^{CFG}(z_t; t,c) - \boldsymbol{\epsilon}\right) \frac{\partial z_t}{\partial \Theta},
\end{equation}
where $\hat{\boldsymbol{\epsilon}}_\phi^{CFG}(z_t; t,c)$ is the noise estimate (Eq.~\ref{eq:p_phi}) under Classifier-Free Guidance (CFG).

\noindent\textbf{Classifier-Free Guidance (CFG).} CFG~\cite{ho2022classifierfree} balances conditional and unconditional predictions through adjustable weights. For a single-condition (text-only) setup, the CFG prediction is:
\begin{equation}\label{eq_text_cond_cfg}
        \hat{\boldsymbol{\epsilon}}_\phi^{CFG}(z_t, t, y) = (1+\omega)\,\hat{\boldsymbol{\epsilon}}_\phi(z_t, t, y) - \omega\,\hat{\boldsymbol{\epsilon}}_\phi(z_t, t, \emptyset),
\end{equation}
where $\omega$ controls the strength of the guidance.

\begin{figure}[!t]
    \centering
    \includegraphics[width=\linewidth]{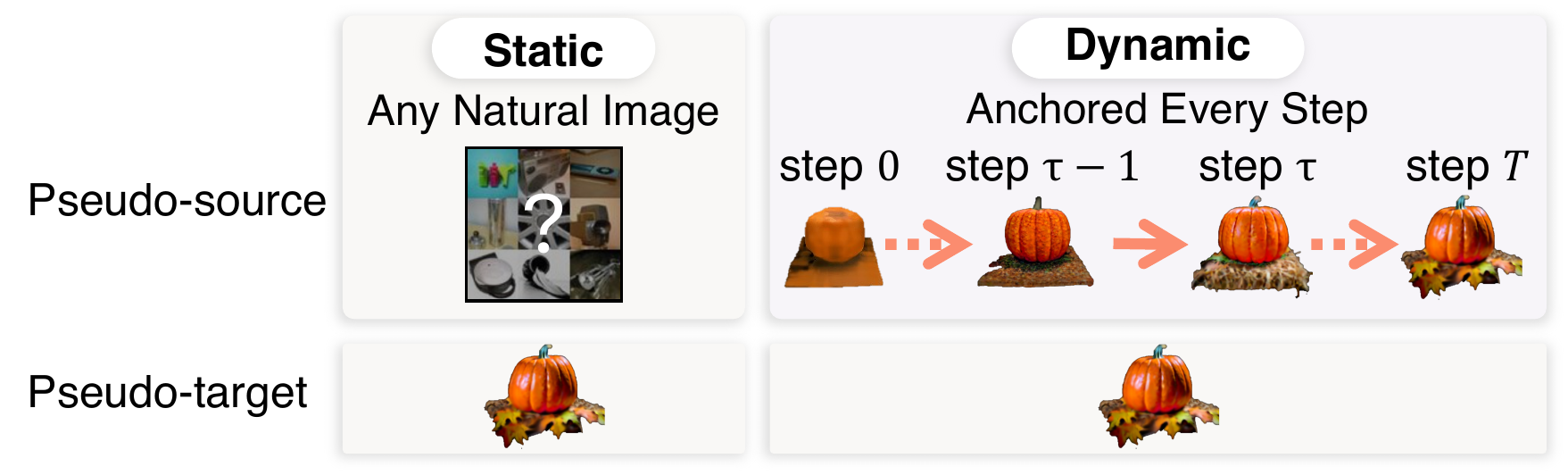}
    \vspace{-20pt}
    \caption{\small\textbf{Static vs. Dynamic Pseudo-Editing.} Static source estimation leverages an unconditional prior, misaligned with the actual 3D state. Instead, the dynamic pseudo-source reflects the evolving 3D renderings, ensuring faithful guidance.}
    \label{fig:static_vs_dynamic}
    \vspace{-10pt}
\end{figure}

\subsection{{The Issue of Static Source Estimation}}

\begin{figure*}[!t]
    \centering
    \includegraphics[width=\linewidth]{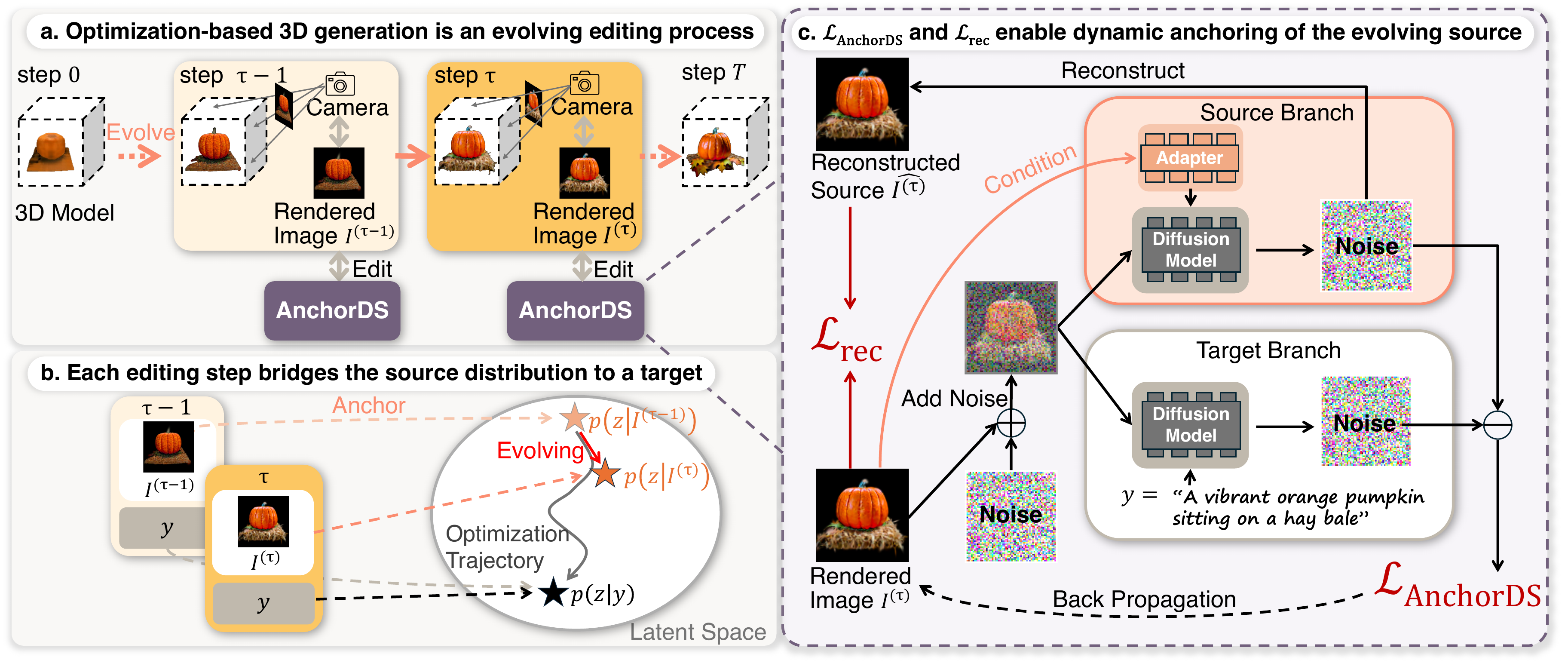}
    \vspace{-20pt}
    \caption{\small\textbf{Overview of Our \ourSDS\ for Text-to-3D Generation.} \textbf{(a)} Optimization-based 3D generation as an evolving editing process where each step refines the 3D model guided by \ourSDS. \textbf{(b)} Each editing step bridges the source distribution to a target distribution through dynamic anchoring in latent space. \textbf{(c)} Technical framework showing how $L_{AnchorDS}$ and $L_{rec}$ enable dynamic source anchoring, with source and target branches processing the evolving 3D content through pretrained diffusion models.}
    \label{fig:framework}
    \vspace{-10pt}
\end{figure*}

We first analyze why static source modeling in SDS causes artifacts (Fig.~\ref{fig:teaser}). Substituting Eq.~\ref{eq_text_cond_cfg} into Eq.~\ref{eq:van_sds} leads to two key terms $m_1$ and $m_2$:
\begin{equation}
\label{eq:sds_decompose_1}
\begin{split}
\!\!\!\! \hat{\boldsymbol{\epsilon}}_\phi^{CFG}(z_t; t, c)\! -\! \boldsymbol{\epsilon} 
    &= (\omega\!-\! 1)\!\cdot\!\underbrace{\bigl(\hat{\boldsymbol{\epsilon}}_\phi(z_t, t, c) -  \hat{\boldsymbol{\epsilon}}_\phi(z_t, t, \emptyset)\bigr)}_{m_1} \\
  \!\! &+  \underbrace{\hat{\boldsymbol{\epsilon}}_\phi(z_t, t, \emptyset)
   - \boldsymbol{\epsilon}}_{m_2}.
\end{split}
\end{equation}
A sufficiently large scaling of $\omega$, \eg, $\omega=100$ will cause $m_1$ to dominate the guidance, while $m_2$ reduces variance {of $\hat{\boldsymbol{\epsilon}}_\phi^{CFG}$}~\cite{mcallister2024sdsbridge,tang2023stable}. Interpreting $m_1$ as a score function:
\begin{equation}\label{eq:m1_interpretation}
\begin{split}
 m_1 %
     = -\sigma_t\Bigl(\nabla_{z_t}\log p(z_t;t,c)-\nabla_{z_t}\log p(z_t;t,\emptyset)\Bigr),
\end{split}
\end{equation}
and we see SDS pushes samples toward the higher-density regions of the conditional distribution $p(z_t;t,c)$ while moving away from the unconditional prior $p(z_t;t,\emptyset)$.

This mirrors 2D image editing (\eg, DDIB~\cite{su2022ddib}, SDEdit~\cite{meng2022sdedit}), where edits aim to find the optimal mapping from a source image to a target image as two distributions. However, SDS approximates the source distribution—which should reflect the current 3D state—using the unconditional prior $p(z_t; t, \emptyset)$. %
The limitation becomes clear when rearranging Eq.~\ref{eq:m1_interpretation} into a pseudo-editing formulation. Specifically, {we invert Eq.~\ref{eq:z_0_to_z_t} to define the pseudo-target and pseudo-source latent reconstructions:}
\begin{equation}
    \begin{split}
        \hat{z}_{t\rightarrow 0}^{\text{target}} &= \frac{1}{\sqrt{\bar{\alpha}_t}} \left( z_t - \sqrt{1-\bar{\alpha}_t}\,\hat{\boldsymbol\epsilon}_\phi(z_t,t,y)\right), \\
\hat{z}_{t\rightarrow 0}^{\text{source}} &= \frac{1}{\sqrt{\bar{\alpha}_t}} \left( z_t - \sqrt{1-\bar{\alpha}_t}\,\hat{\boldsymbol\epsilon}_\phi(z_t,t,\emptyset)\right).
    \end{split}
\end{equation}
Then, the core update term simplifies elegantly as:%
\begin{equation}
m_1 = \eta \left(\hat{z}_{t\rightarrow 0}^{\text{target}} - \hat{z}_{t\rightarrow 0}^{\text{source}}\right),
\quad\text{where}\quad \eta = \frac{\sqrt{\bar{\alpha}_t}}{\sqrt{1 - \bar{\alpha}_t}}.
\end{equation}  
Examining $\hat{z}_{t\rightarrow 0}^{\text{source}}$, we reveal that it loses critical information about the original rendering: (1) $z_t$ is a noise-corrupted version that discards semantic content, while (2) $\hat{\boldsymbol\epsilon}_\phi(z_t,t,\emptyset) = -\sigma_t\nabla_{z_t}\log p(z_t;t,\emptyset)$ pushes toward an unconditional image prior that averages over diverse natural images. Crucially, neither term encodes the evolving 3D asset's current state, forcing $\hat z_{t\rightarrow 0}^{\text{source}}$ to inherit this static, averaged characteristic rather than reflecting the dynamic source distribution. %
As a result, it fails to capture the semantics of intermediate renderings. The resulting update directions become inconsistent with the 3D asset's actual appearance, causing over-smoothing and cross-view inconsistency in Fig.~\ref{fig:teaser}.

Rather than static, we argue that the source should evolve with the 3D model (Fig.~\ref{fig:static_vs_dynamic}). We therefore reinterpret the text-to-3D optimization as a \textit{dynamic edit}: at each step, we refine the 3D asset by { (1) preserving favorable attributes (\eg, structural semantics consistent with current state) via faithful source modeling, and (2) correcting undesirable features (\eg, deviations from target distribution) through prompt alignment. }This resolves the inconsistency in Fig.~\ref{fig:teaser} by anchoring updates to the actual state rather than a static prior.

\section{Method}

\subsection{\!\!{Dynamic Evolution of the Source Distribution}\!\!\!\!\!}\label{sec:dynamic_evolution}

We formalize a critical insight regarding SDS-based text-to-3D generation: the source image distribution $P_\theta(x)$ evolves dynamically throughout optimization rather than remaining static, as visualized in Fig.~\ref{fig:framework}b. Mathematically, we reframe SDS optimization as a progressive transformation process:
\begin{equation}
P^{(0)}_\theta(x) \xrightarrow{\!\nabla_{\Theta}\mathcal{L}_{\rm SDS}} P^{(1)}_\theta(x) \xrightarrow{} \cdots \xrightarrow{} P^{(T)}_\theta(x),
\end{equation}
where $P^{(\tau)}_\theta(x)$ denotes the time-varying source distribution at optimization step $\tau$. Initially, $P^{(0)}_\theta(x)$ resembles a diffuse unconditional prior when the 3D model $\Theta$ is randomly initialized. At each iteration, the gradient update $\nabla_{\Theta}\mathcal{L}_{\rm SDS}$ transports probability mass such that:
\begin{equation}
P^{(\tau+1)}_\theta(x) = \mathcal{T}\left(P^{(\tau)}_\theta(x), \nabla_{\Theta}\mathcal{L}_{\rm SDS}(z,y, \hat{\boldsymbol\epsilon}_t)\right),
\end{equation}
where $\mathcal{T}(\cdot)$ represents the distributional shift operator. This process progressively morphs $P^{(\tau)}_\theta(x)$ toward the target distribution $P_{\rm target}(x\,|\,y)$ defined by the text prompt $y$. The continuous distributional evolution stems from the mass transport interpretation of score-based dynamics~\cite{de2021diffusion}, establishing this formulation's generality across score-distillation-based 3D generation methods.

Next, we introduce our \ourSDS\ %
to ensure accurate source distribution estimation at every optimization step.

\vspace{-3pt}
\subsection{\!\!\!\!\!{Score Distillation via Dynamic Source Anchoring}\!\!\!\!\!\!}\label{sec:analysis}

AnchorDS dynamically anchors the score guidance at the evolving 3D state through image conditioning. Formally, at optimization step $\tau$ with rendered view $I^{(\tau)}$, we compute the guidance gradient as:
\begin{equation}\label{eq:sds_our}
    g_t^{(\tau)} = \hat{\boldsymbol{\epsilon}}_\phi(z_t; t, y) - \hat{\boldsymbol{\epsilon}}_\phi(z_t; t, \emptyset, I^{(\tau)}),
\end{equation}
where $\hat{\boldsymbol\epsilon}_\phi(z_t;t,y)$ targets the text-conditioned distribution, while $\hat{\boldsymbol\epsilon}_\phi(z_t;t,\emptyset,I^{(\tau)})$ anchors the current source distribution $P^{(\tau)}_\theta(x)$. This differential formulation directs updates from the current state toward the target distribution.

Essentially, incorporating an image condition $I^{(\tau)}$ recasts the problem into a dual-conditioned latent space. This preserves the text-conditioned target's effectiveness while leveraging image conditions to anchor the source distribution. Specifically, $I^{(\tau)}$ anchors the current source by providing structural grounding without content constraints, enabling contextual editing rather than output restriction.
It yields two principal advantages:
(1) %
Conditioning with $I^{(\tau)}$ at continuous $\tau$ maintains alignment between guidance and 3D state, mitigating drift and oversmoothing.
(2) %
Source anchoring requires only a single additional U-Net forward pass per iteration, processed in parallel with the original pass, thus maintaining identical runtime to standard SDS.

AnchorDS bridges 2D editing principles with 3D generation through a key insight: pretrained diffusion models inherently possess image inversion capabilities within their conditional architecture. While 2D editing requires explicit inversion techniques, AnchorDS elegantly leverages this intrinsic property—directly utilizing the model's natural ability to map images to latent distributions, achieving precise source anchoring without auxiliary inversion costs.

\vspace{-6.5pt}
\subsection{Source Anchoring via Image Conditioning}\label{sec:img_cond_src_est}
\vspace{-1pt}

The effectiveness of \ourSDS\ depends critically on an accurate estimation of the current source distribution via image-conditioned diffusion models. We denote by $I^{(\tau)} = R(\mathcal{X}_{\text{3D}}^{(\tau)})$ the rendered image %
of the 3D representation $\mathcal{X}_{\text{3D}}^{(\tau)}$.

\noindent\textbf{Choice of Image Condition.} In general, one could apply a preprocessing function $E(\cdot)$ to obtain a conditioning signal $\bar I^{(\tau)} = E(I^{(\tau)})$ (\eg, converting $I^{(\tau)}$ to a normal map, depth map, etc.) %
The conditioning signal must retain essential structural and semantic information from the 3D rendering while eliminating irrelevant noise. We initially consider that %
the identity mapping $E(I^{(\tau)}) = I^{(\tau)}$ is particularly effective as it preserves maximal information about the current state. This aligns with evidence~\cite{kadosh2025tight} that image-conditioned diffusion models learn invertible mappings between images and noise—crucial for source estimation. Meanwhile, we find that our method remains compatible with alternative signals (\eg, normal maps) provided that they sufficiently capture the 3D content's core attributes. 

\noindent\textbf{Pseudo-Source Reconstruction.}
With the image-conditioned model in place, we can obtain an explicit estimate of the current source image distribution at any diffusion step. 
Given a noisy latent $z_t$ (at noise level $t$) that was produced from the current image $I^{(\tau)}$, the model's image conditioned prediction $\hat{\boldsymbol\epsilon}_\phi(z_t;t,\emptyset,I^{(\tau)})$ allows us to reconstruct a pseudo-reconstructed source image:
\begin{equation}\label{eq:one_step_zt0_img_cond}
\hat{z}^{\text{anchored},(\tau)}_{t\rightarrow 0} = \frac{1}{\sqrt{\bar{\alpha}_t}}\Big( z_t - \sqrt{1-\bar{\alpha}_t}\; \hat{\boldsymbol\epsilon}_\phi(z_t; t,\emptyset, I^{(\tau)})\Big),
\end{equation}
which is the model's one-step estimate of the denoised latent at timestep $t$ – essentially a guess of the original image $I^{(\tau)}$ (latent) before noise was added.
With an image decoder $\varepsilon(\cdot)$, we can now explicitly evaluate reconstruction accuracy as:
\begin{equation}\label{eq:rec_loss}
\mathcal{L}_{\text{rec}} = \big\|\varepsilon(\hat{z}^{\text{anchored},(\tau)}_{t\rightarrow 0}) - I^{(\tau)}\big\|^2_2.
\end{equation}
This reconstruction not only provides a direct metric for source estimation quality but also enables two complementary mechanisms: (1) filtering out unstable predictions, and (2) fine-tuning the image adapter to better align with rendered images—both crucial for stable and accurate 3D optimization. We implement two strategies discussed below.

\begin{figure}[!t]
\centering\hspace{-0.49em}\includegraphics[width=1.02\linewidth]
{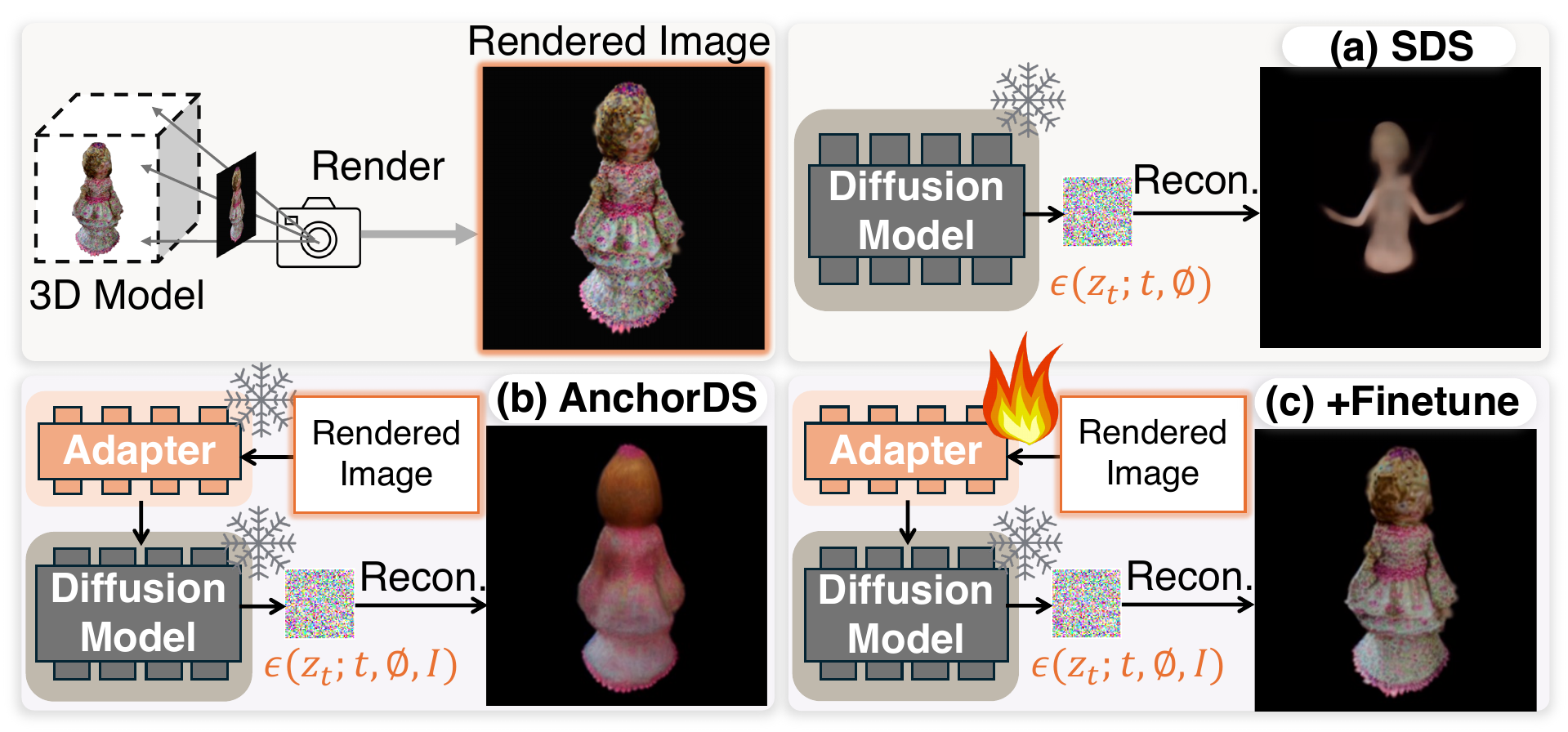}
    \vspace{-20pt}
    \caption{\small \textbf{Source Estimation Strategies.} (a) Vanilla SDS poorly reconstructs the source. (b) \ourSDS, conditioned on the current rendering, recovers geometry and color faithfully. (c) Our fine-tuning strategy achieves an accurate match to the source image.
}
    \label{fig:src_est_compare}
    \vspace{-15pt}
\end{figure}

\noindent\textbf{Filtering.}
A straightforward way to employ the objective (Eq.~\ref{eq:rec_loss}) is to apply a threshold-based filter to exclude unreliable source estimations:
\vspace{-2pt}
\begin{equation}\label{eq:filter}
\begin{split}
        \mathcal M_\text{Filter} = \begin{cases} 1 & \text{if } \mathcal{L}_{\text{rec}}<\gamma \\ 0 & \text{otherwise} \end{cases}\\
    \mathcal L_\text{AnchorDS} = \mathcal M_\text{Filter} \cdot \mathcal L_\text{AnchorDS}.
\end{split}
\end{equation}
Here, $\mathcal{L}_{\text{rec}}$ acts as a stabilizer, filtering out any spurious deviations in the anchored source prediction. Empirically, this translates to improved stability (no sudden jarring updates) and better preservation of existing content.

\noindent\textbf{Finetuning Image Adapter.}
Another approach is to apply $\mathcal L_\text{rec}$ to fine-tune the image adapter. Although pretrained 2D models are powerful enough to model the real-world image distribution, there remains a gap between the real-world and the synthesized rendered image distribution. Intuitively, we aim to let 2D models ``see'' the real data. We address this through lightweight fine-tuning of the image adapter using $\mathcal{L}_{\text{rec}}$. This fine-tuning is minimal—only unfreezing %
one single layer of the image adapter is sufficient (increasing optimization time from $\sim$25 minutes to $\sim$30 minutes using 3DGS pipeline~\cite{yi2023gaussiandreamer}), while significantly improving source estimation accuracy (see Fig.~\ref{fig:src_est_compare}c).

Minimizing $\mathcal{L}_{\text{rec}}$ forces $\hat{\boldsymbol\epsilon}_\phi(z_t;t,\emptyset,I^{(\tau)})$ to become consistent with the current image $I^{(\tau)}$. %
Note $\hat{\boldsymbol\epsilon}_\phi(z_t;t, \emptyset,I^{(\tau)})$ is \emph{not} trained to match sampled noise; Instead, its role is to ``invert'' the current image into latent space. Eq.~\ref{eq:one_step_zt0_img_cond} simply uses this predicted noise to retrieve the model's internal guess of the clean image $I^{(\tau)}$, and $\mathcal{L}_{\text{rec}}$ ties that guess further. 
With this enhanced source estimate term, guidance $g_t^{(\tau)}$ (Eq.~\ref{eq:sds_our})  accurately reflects the difference between distributions of images that ensembles $I^{(\tau)}$ and those that fulfill $y$. 

\vspace{-6pt}
\subsection{Implementation}\label{sec:implementation}

\noindent\textbf{Choice of Diffusion Model.}
To incorporate $\bar I^{(\tau)}$ as a condition, we leverage existing pre-trained dual-conditional diffusion models. A straightforward option is an image-to-image diffusion model such as IP2P~\cite{brooks_instructpix2pix_2023}. However, IP2P is fine-tuned for image editing and we observed it gives inconsistent guidance under the high CFG weights required for SDS; in practice, using IP2P led to unstable and desaturated results for text-to-3D, especially at large CFG scales. 
ControlNet~\cite{zhang2023controlnet} processes derived maps (\eg, normal maps, sketches) but lacks direct training on raw images. Conversely, IP-Adapter~\cite{ye2023ipadapter} conditions Stable Diffusion (1.5) on unaltered images via an auxiliary latent, preserving the model's expressive power without constraining image content. Crucially, our framework generalizes across adapters.%
We adopt IP-Adapter for primary experiments due to its direct image conditioning, while adopting ControlNet shows comparative results (detailed in Sec.~\ref{sec:experiments}). %

\vspace{-5pt}
\subsubsection{Pipeline.} Given a text prompt $y$ and an optional initial 3D representation (which could be random or from a rough generator), our pipeline operates iteratively as illustrated in Fig.~\ref{fig:framework}c. At each optimization step $\tau$, we render the current 3D model from a random viewpoint to obtain image $I^{(\tau)}$, then encode it into the diffusion latent space and add noise at a random timestep $t$ to produce $z_t$. We then query the diffusion model twice: once with text condition $y$ to obtain the target prediction $\hat{\boldsymbol\epsilon}_\phi(z_t; t, y)$, and once with an empty text and $I^{(\tau)}$ to obtain the source prediction $\hat{\boldsymbol\epsilon}_\phi(z_t; t, \emptyset, I^{(\tau)})$. The obtained AnchorDS guidance (Eq.~\ref{eq:sds_our}) is then backpropagated through the rendering pipeline to update the 3D parameters. 
For the fine-tuning variant, we periodically update the unfrozen layer of the image adapter using $\mathcal{L}_{\text{rec}}$. %

\vspace{-8pt}
\section{Experiments}\label{sec:experiments}

\begin{figure*}[t]
    \centering
    \includegraphics[width=0.95\linewidth]{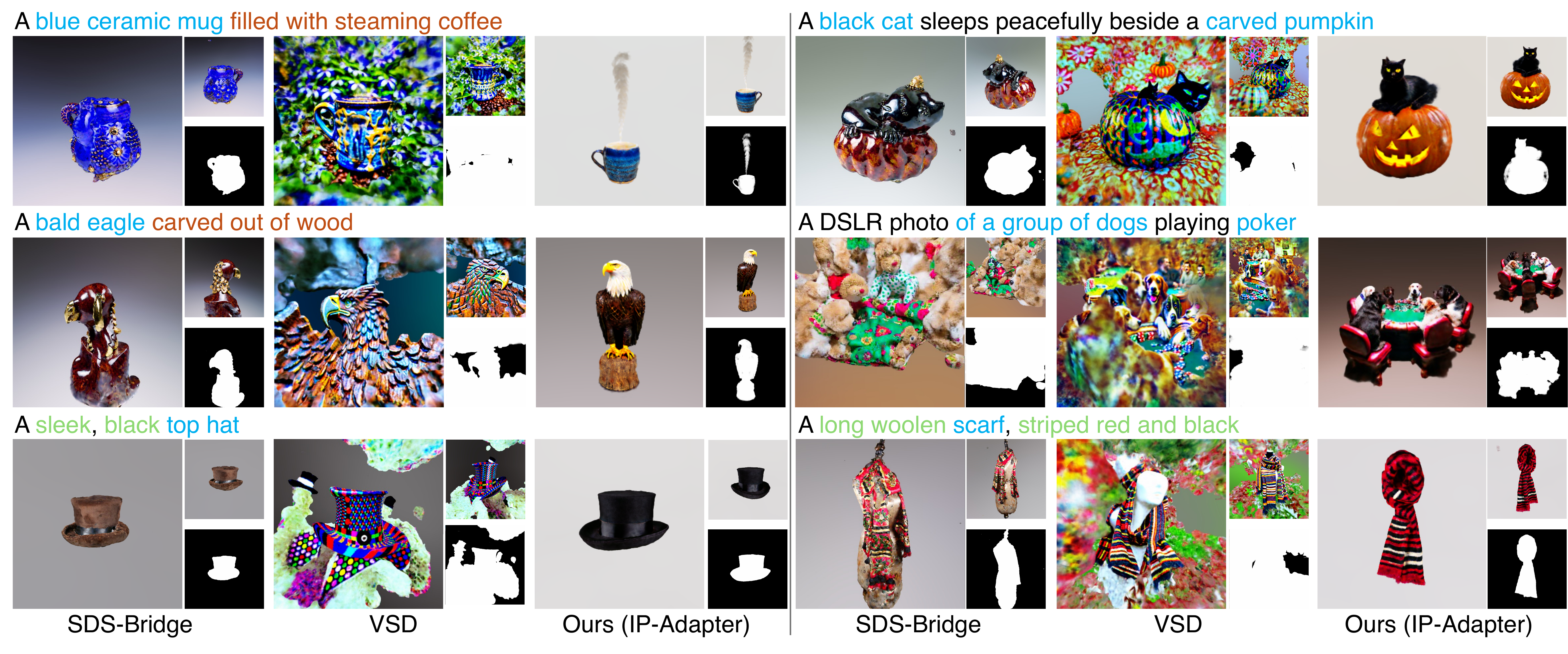}
    \vspace{-12pt}
    \caption{\small \textbf{Qualitative Comparison to Existing Methods on SD 1.5} on complex text prompts, including 
(1) \textcolor{myblue}{objects} with \textcolor{myorange}{rich or mixed semantics}; 
(2) \textit{multiple} distinct \textcolor{myblue}{objects}; 
(3) \textcolor{myblue}{objects} with \textcolor{mygreen}{fine-grained details}.
    SDS-Bridge produces biased material textures and inaccurate semantics, likely due to new biases introduced by the handcrafted prompt. VSD fails to recover coherent structures and exhibits exaggerated, unrealistic colours. %
    In contrast, our method consistently generates semantically faithful and structurally accurate results. }
    \label{fig:compare_sb_vsd}
    \vspace{-10pt}
\end{figure*}

\noindent\textbf{Experimental Setup.}
We aim to evaluate the effectiveness of \ourSDS\ in addressing the limitations of SDS guidance, specifically source distribution mismatch and semantic over-smoothing. We compare against vanilla SDS~\cite{poole_dreamfusion_2022} and two representative methods tackling similar issues: SDS-Bridge~\cite{mcallister2024sdsbridge} and ProlificDreamer (VSD)\cite{wang_prolificdreamer_2023}.
To validate robustness, we test across both 3D Gaussian Splatting (3DGS)\cite{yi2023gaussiandreamer} and NeRF-based pipelines.

\noindent\textbf{Evaluation Metrics.}
Following \cite{mcallister2024sdsbridge,lee2024dreamflow,dong2024coin3d}, we employ CLIP similarity~\cite{radford2021clip} between text prompts and rendered images to assess generation alignment. Additionally, we adopt T$^3$Bench~\cite{he2023t3bench}'s quality metric, 
evaluating the 3D visual quality using pre-trained language and visual models. Since the target estimation term in Eq.~\ref{eq:m1_interpretation} remains unchanged, text alignment is maintained by design, letting us focus primarily on quality improvements.

\vspace{-5pt}
\subsection{Comparison with State-of-the-Art}

\setlength{\tabcolsep}{12pt}
\begin{table}[tb!]
\vspace{-6pt}
  \centering
  \small
  \caption{\small \textbf{Quantitative Comparison.} Q1–Q3 report averaged ranking of each method.}
  \vspace{-10pt}
  \renewcommand{\arraystretch}{1.1}

  \scalebox{0.7}{%
        \begin{tabular}{l|c|c|ccc}
          \toprule
          \textbf{Method} & \textbf{Base Model\,} & \textbf{CLIP}~$\uparrow$ & \textbf{Q1}~$\downarrow$ & \textbf{Q2}~$\downarrow$ & \textbf{Q3}~$\downarrow$ \\
          \midrule
          VSD               & \multirow{2}{*}{SD 2.1} & 0.352          & 1.84 & 1.85 & 1.79 \\
          Ours (ControlNet) &                         & \textbf{0.369} & \textbf{1.16} & \textbf{1.15} & \textbf{1.21} \\
          \midrule
          VSD               & \multirow{3}{*}{SD 1.5} & 0.281          & 1.99 & 2.00 & 2.08 \\
          SDS-Bridge        &                         & 0.233          & 2.38 & 2.35 & 2.29 \\
          Ours (IP-Adapter) &                         & \textbf{0.334} & \textbf{1.63} & \textbf{1.66} & \textbf{1.63} \\
          \bottomrule
        \end{tabular}
        \vspace{5pt}
    }
    \scalebox{0.7}{
        \parbox{1.5\linewidth}{
        \raggedright
        Q1: Which one has the best 3D consistency?\\
        Q2: Which one shows accurately what the text describes?\\
        Q3: Which one looks most realistic and natural?}
    }%

  \label{tab:quant-comparison}
  \vspace{-16pt}
\end{table}

To systematically evaluate SDS's limitations, we curate 50 complex prompts from previous works~\cite{he2023t3bench,poole_dreamfusion_2022}, covering three challenging categories: fine-grained details, rich/mixed semantics, and multiple-object compositions. This evaluation scale aligns with established practices~\cite{mcallister2024sdsbridge,liang2024luciddreamer,lee2024dreamflow,dong2024coin3d}.
We adopt the NeRF-based pipeline~\cite{wang_prolificdreamer_2023} across all experiments. While VSD demonstrates sensitivity to base models and typically performs reasonably only on Stable Diffusion (SD) 2.1, we evaluate on both SD 1.5 and SD 2.1. Our method (AnchorDS with Finetuning) uses IP-Adapter on SD 1.5 and ControlNet on SD 2.1 as image conditioners.

\noindent\textbf{Human Evaluation.} We also conduct a comprehensive human preference evaluation on Amazon Mechanical Turk to assess methods across multiple dimensions. The evaluation comprises 20 batches covering our 50 complex prompts evaluated by 912 unique participants. For each comparison, evaluators rank all methods according to three questions: 3D consistency, text alignment, and visual quality, respectively.

\begin{figure*}[tbh]
    \centering
    \includegraphics[width=0.95\linewidth]{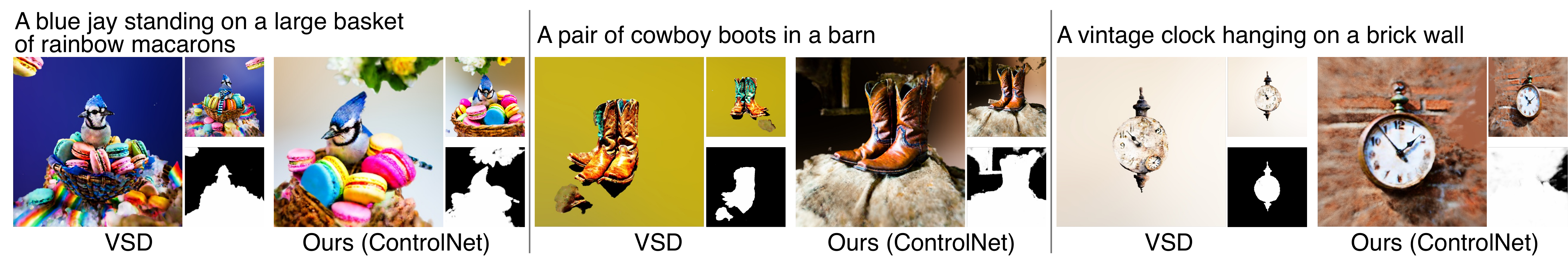}
    \vspace{-13pt}
    \caption{\small \textbf{Qualitative Comparison to Existing Methods on SD 2.1.}
    Our method surpasses VSD, consistently capturing complex semantics (even challenging elements like barns and brick walls) while ensuring superior 3D structural accuracy. }
    \label{fig:compare_sd21}
        \vspace{-10pt}
\end{figure*}

\begin{figure*}[!t]
    \centering
    \includegraphics[width=0.95\linewidth]{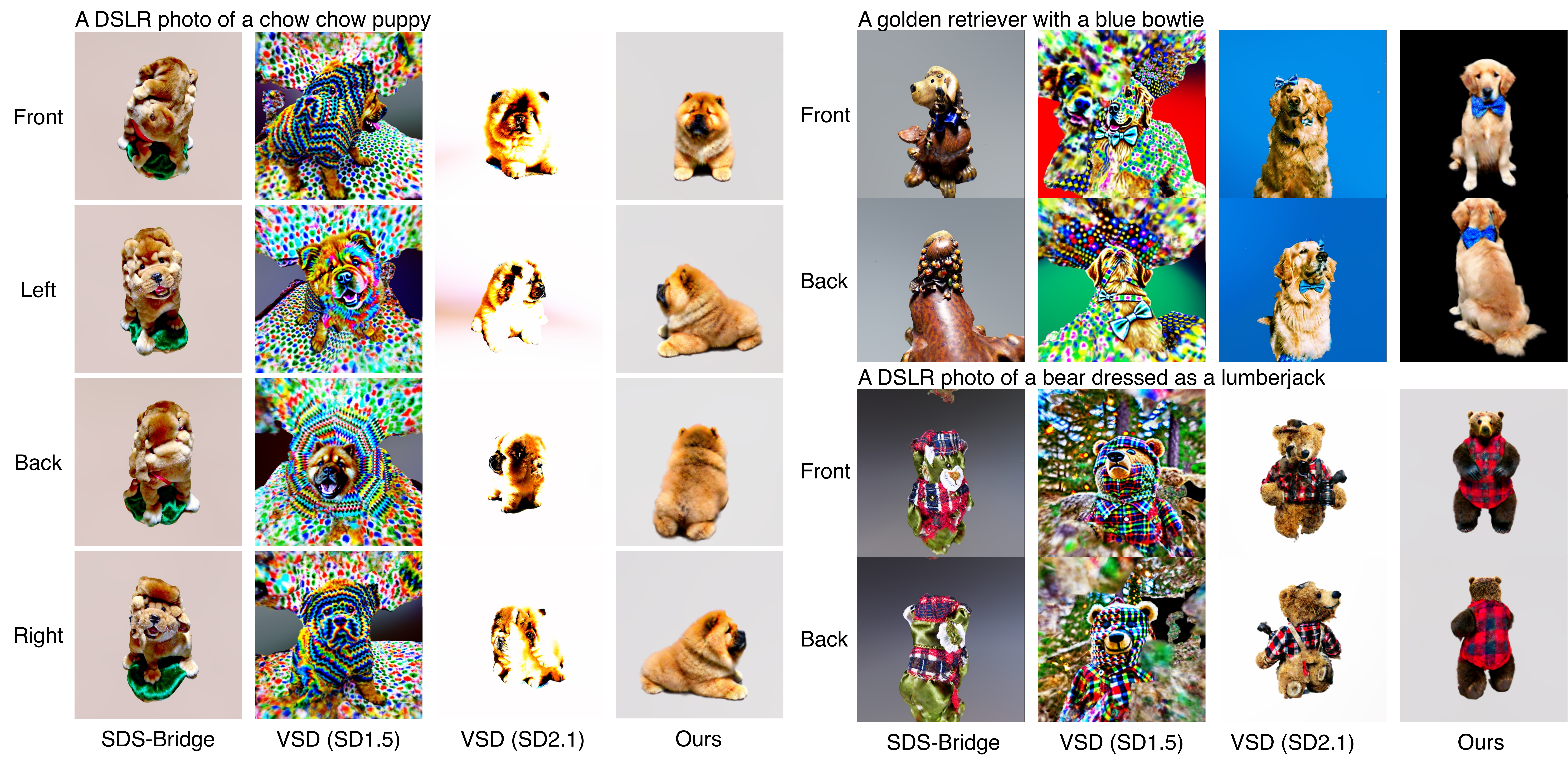}
    \vspace{-13pt}
    \caption{\small \textbf{Multi-view Comparison.} Ours demonstrates significantly better quality and multi-view consistency, despite not incorporating common 3D consistency strategies such as Perp-Neg~\cite{armandpour2023perpneg}. In contrast, although VSD introduces view direction as an additional condition, it still suffers from a severe Janus problem. SDS-Bridge fails to maintain consistent object semantics.}%
    \label{fig:compare_sb_vsd_multiview}
    \vspace{-8pt}
\end{figure*}

\noindent\textbf{Results.} Table~\ref{tab:quant-comparison} shows our method consistently outperforms all methods in CLIP similarity and human preference evaluations. %
Qualitative comparisons in Fig.~\ref{fig:compare_sb_vsd} for SD 1.5 as base model, and Fig.~\ref{fig:compare_sd21} for SD 2.1 as base model, and a multiview structural consistency comparison Fig.~\ref{fig:compare_sb_vsd_multiview} also demonstrate our consistent generation of semantically faithful and structurally accurate results, while SDS-Bridge produces artifacts introduced by biased negative prompts, and VSD fails to accurately model the image distributions. Please refer to the Appendix for theoretical discussions.

\vspace{-8pt}
\subsection{Baseline Comparison}
\noindent\textbf{SDS Baselines.}
We comprehensively evaluate on T$^3$Bench~\cite{he2023t3bench}, a comprehensive benchmark containing 300 prompts across single object generation, object with surrounding generation, and multiple object generation categories. We compare against both DreamFusion-SDS and GaussianDreamer-SDS baselines. 
To ensure fair evaluation, we maintain all pipeline components constant except for the score guidance, using the state-of-the-art GaussianDreamer~\cite{yi2023gaussiandreamer} framework as our baseline. All methods are built upon SD 1.5.%

\begin{figure*}[!t]
    \centering
    \includegraphics[width=0.95\linewidth]{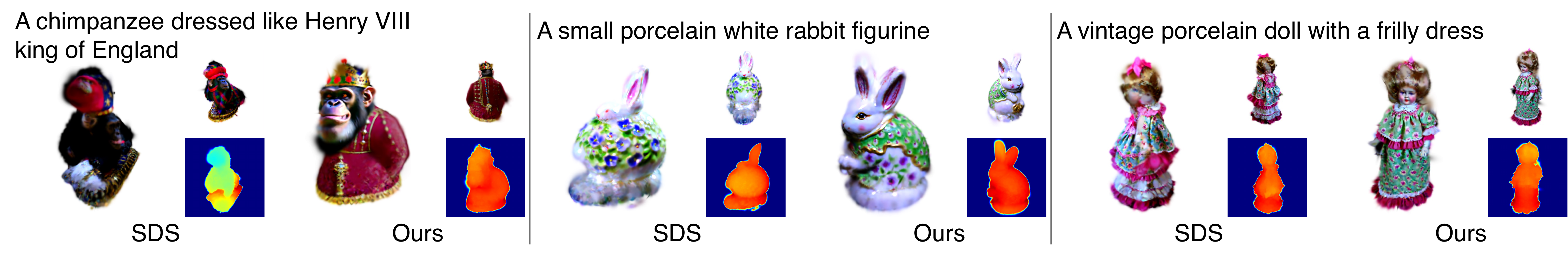}
    \vspace{-13pt}
    \caption{\small \textbf{Qualitative Comparison with Baseline (SDS).} Ours shows superior visual quality in finer-grained detail (\eg, doll's face). For prompts describing a single object with complex semantic details (\eg, porcelain white rabbit figurine), SDS tends to compress or mix parts of the information, whereas ours successfully preserves the full range of semantic attributes. For prompts involving multiple objects (\eg, chimpanzee and dresses), SDS appears mixed and blurry, while ours accurately separates distinct objects.}
    \label{fig:compare_sds}
        \vspace{-14pt}
\end{figure*}

\noindent\textbf{Comparison with SDS.} Quantitative results in Table~\ref{tab:t3bench-quality} demonstrate that we consistently surpass SDS in all prompt categories. Our method achieves superior visual quality as illustrated in Fig.~\ref{fig:compare_sds}, with notable improvements in finer-grained detail, the preservation of complex semantic attributes, and the separation of multiple objects. 
The improvements align with our theoretical framework: Being aware of the evolving 3D state, AnchorDS generates along coherent semantic paths without mixing disparate attributes.

\setlength{\tabcolsep}{13.5pt}
\begin{table}[tb]
        \vspace{-6pt}
    \tiny
    \centering
    \caption{\small\textbf{Quantitative Comparison with Baseline on T$^3$Bench Quality Metric.} AnchorDS incorporates image conditioning via IP-Adapter or ControlNet. Filter and Finetune represent alternative strategies for enhanced source distribution estimation.}
    \label{tab:t3bench-quality}
    \vspace{-10pt}
   \renewcommand{\arraystretch}{1.1}
    \scalebox{0.9}{
        \begin{tabular}{l|cccc}
        \toprule
        \textbf{Method} & \textbf{All}$\uparrow$ & \textbf{Single}$\uparrow$ & \textbf{Surr}$\uparrow$ & \textbf{Multi}$\uparrow$ \\
        \midrule
        SDS (DreamFusion)                & 20.5 & 24.9 & 19.3 & 17.3  \\
        SDS (GaussianDreamer)                & 29.7 & 42.3 & 26.1 & 20.6  \\\midrule \midrule
        AnchorDS (IP-Adapter)        & 30.7 & 43.0 & 24.8 & 24.5  \\     
        + Filter   & {32.8} & {44.1} & {27.9} & \textbf{26.5}  \\  
        + Finetune & \textbf{33.3} & \underline{45.3} & \underline{29.0} & \underline{25.7}  \\\midrule
        {AnchorDS (ControlNet)} & {30.8} & {43.9} & {27.2} & {21.3}  \\
        {+ Filter} & \underline{33.2} & \textbf{46.1} & \textbf{29.4} & 24.0  \\
        {+ Finetune} & {32.9} & {45.0} & {28.6} &  25.2 \\
        \bottomrule
        \end{tabular}
        }
        \vspace{-15pt}
\end{table}

    \vspace{-8pt}
\subsection{Ablation Studies}\label{sec:ablation}
 Quantitative ablation results for our key components are shown in Table~\ref{tab:t3bench-quality}. AnchorDS consistently outperforms vanilla SDS through source anchoring via image conditioning. Our Filtering and Finetuning strategies provide complementary approaches for enhancing source estimation accuracy, with Finetuning achieving optimal performance. Please refer to Appendix for more ablations across prior models.

\section{Conclusion}
We demonstrate that the source distribution dynamically evolves during text-to-3D optimization—a fundamental property that has been largely overlooked by existing methods. While prior efforts have focused on reducing trajectory estimation error or improving the guidance prior, they continue to exhibit critical issues such as semantic over-smoothing and multi-view inconsistency, due to inaccurate modeling of the evolving source. To address this, we introduce \ourSDS, which anchors the dynamic source distribution by casting the problem into a dual-conditioned latent space and conditioning on rendering images. Experimental results confirm our method effectively mitigates these issues and achieves superior fidelity and visual quality.

\FloatBarrier
\bibliography{egbib}

\clearpage

\appendix

\setcounter{secnumdepth}{0}

\section*{AnchorDS: Anchoring Dynamic Sources for Semantically Consistent Text-to-3D Generation \\ \textbf{--Appendix--}}

\section{Implementation Details}

\noindent\textbf{3D Representation.}
We evaluate \ourSDS\ using both 3D Gaussian Splatting (3DGS)~\cite{yi2023gaussiandreamer} and NeRF as 3D representations to demonstrate robustness across different representation types. These representations exhibit different sensitivities to SDS guidance: 3DGS, being point-cloud-like, is highly sensitive to gradient flickering and often requires stabilization strategies like gradient clipping and hierarchical optimization. NeRF, while more stable, requires significantly more optimization steps to converge.
For 3DGS experiments, we follow common practice by initializing with simple text-to-3D point clouds from Shap-E~\cite{jun2023shap_e}, as 3DGS cannot converge without reasonable initialization. This adds minimal computational cost due to the explicit nature of the representation.
Importantly, unlike SDS-Bridge which requires SDS-guided rough initialization to prevent source estimate deviation, \ourSDS\ does not require such initialization. Our image-conditioned anchoring mechanism accurately captures the source distribution from the beginning, allowing direct optimization without preliminary SDS stages for NeRF experiments.

\noindent\textbf{AnchorDS Guidance Formula.}
Adding back the variance reduction term $m_2$ in Eq. 6 in the main paper, our final AnchorDS guidance is:
\begin{equation}\begin{split}
        \mathcal L_{AnchorDS} =\, &g_t^{(\tau)} + m_2 \\
        =\, &\hat{\boldsymbol{\epsilon}}_\phi(z_t; t, y) - \hat{\boldsymbol{\epsilon}}_\phi(z_t; t, \emptyset, I^{(\tau)}) \\
        &+ \hat{\boldsymbol{\epsilon}}_\phi(z_t, t, \emptyset)
   - \boldsymbol{\epsilon}.
\end{split}
\end{equation}
Negative prompts $y_{\text{neg}}$ may replace $\emptyset$ for more informative anchoring.

\noindent\textbf{Hardware \& Training Setup.}
All experiments are conducted on a single NVIDIA A40 GPU (48 GB). %
The threshold for the Filtering strategy is $\gamma=0.03$, and the image adapter fine-tuning strategy adopts a learning rate of $1\times10^{-4}$. All remaining hyper‑parameters mirror those of GaussianDreamer~\cite{yi2023gaussiandreamer} for the 3DGS pipeline and ProlificDreamer~\cite{wang_prolificdreamer_2023} for the NeRF pipeline.

To clarify the practical cost of our dynamic anchoring, we report wall-clock runtimes in Table~\ref{tab:runtime}. All runs are measured per text prompt. The additional image-conditioning pass in \ourSDS\ is executed in parallel with the original diffusion pass and thus incurs negligible overhead compared to SDS-based baselines. Optional filtering and lightweight adapter fine-tuning slightly increase runtime but remain a minor fraction of the total optimization.

\begin{table}[t]
\centering
\scriptsize
\caption{\small \textbf{Runtime comparison.} Wall-clock runtimes per text prompt on a single NVIDIA A40 GPU. \ourSDS\ matches the cost of SDS-based baselines; the optional filtering (Filter) and adapter fine-tuning (FT) introduce only marginal overhead while improving robustness and fidelity.}
\scalebox{0.85}{
    \begin{tabular}{lccc}
    \toprule
    Method & 3D Representation & Runtime / prompt \\
    \midrule
    GaussianDreamer & 3DGS & 25 min \\
    \ourSDS\ (ours) & 3DGS & 25 min \\
    \ourSDS\ (ours) + Filter + FT & 3DGS & 30 min \\
    \midrule
    ProlificDreamer & NeRF & 3.5 h \\
    \ourSDS\ (ours) & NeRF & 3.5 h \\
    \ourSDS\ (ours) + Filter + FT & NeRF & 4.0 h \\
    \bottomrule
    \end{tabular}
}
\label{tab:runtime}
\end{table}

\section{More Ablation Studies}

\begin{figure*}[tbh]
    \centering
    \includegraphics[width=\linewidth]{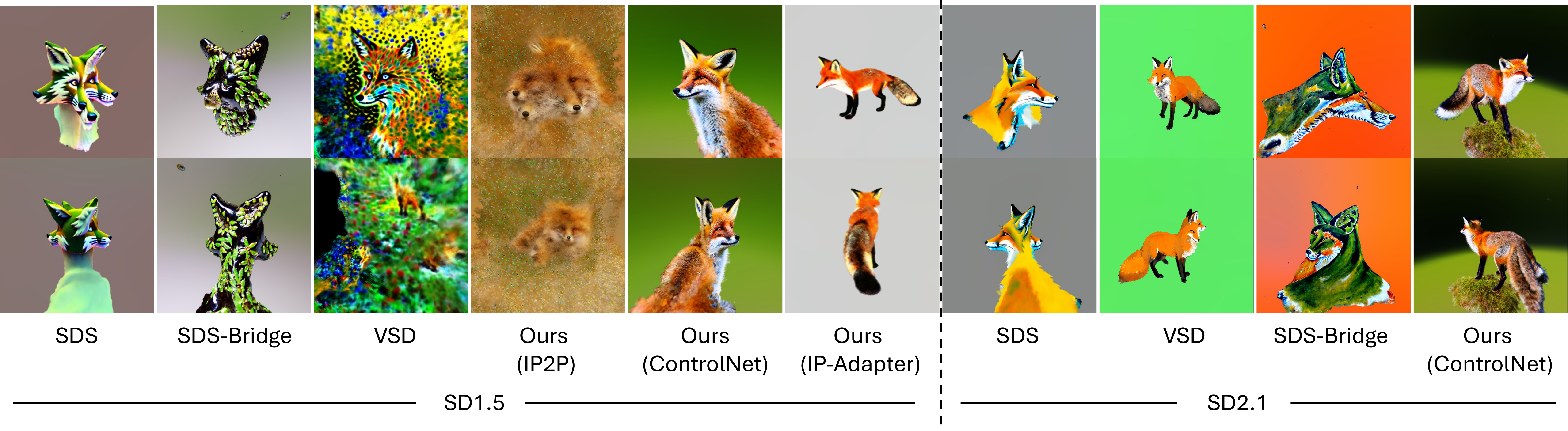}
    \caption{\textbf{Ablation studies across diffusion models and conditioning adapters.} Our AnchorDS variants consistently outperform existing methods (SDS, SDS-Bridge, VSD) across all configurations. Notably, AnchorDS with IP-Adapter on SD 1.5 achieves competitive quality with VSD on SD 2.1, while AnchorDS with ControlNet on SD 2.1 produces the most photorealistic results. Among conditioning adapters, IP-Adapter demonstrates superior 3D consistency compared to IP2P and ControlNet alternatives.}
    \label{fig:ablation_controlnet}
\end{figure*}

To validate the robustness and generalizability of our approach across different prior model configurations, we conduct ablation studies examining the impact of various 2D diffusion models and image conditioning adapters. Quantitative results shown in Table 2 in the main paper also validate that AnchorDS is robust against the selection of different image conditionings, using both ControlNet with normal map and IP-Adapter with the identity image as image conditions achieve better results compared with the baseline.
In Fig.~\ref{fig:ablation_controlnet}, we show the qualitative evaluation of AnchorDS using SD 1.5 with three image conditioners—InstructPix2Pix~\cite{brooks_instructpix2pix_2023} (IP2P), IP-Adapter~\cite{ye2023ipadapter}, and ControlNet-NormalBae~\cite{zhang2023controlnet}. With SD 2.1, we adopt ControlNet-NormalBae conditioning. 

\noindent\textbf{Cross-Model Performance:} Despite SD 1.5's more limited capabilities compared to SD 2.1, our method with IP-Adapter on SD 1.5 achieves competitive results with VSD running on the significantly more powerful SD 2.1 base model. When combined with SD 2.1 and ControlNet, our approach generates the most photorealistic textures among all evaluated configurations. 

\noindent\textbf{Comparison with Existing Methods:} Across all tested configurations, AnchorDS variants consistently outperform existing methods including vanilla SDS, SDS-Bridge, and VSD, demonstrating the fundamental effectiveness of our dynamic source anchoring strategy regardless of the underlying architecture. 

\noindent\textbf{Adapter Comparison:} Among our conditioning variations, both IP-Adapter and ControlNet produce high-quality and 3D-consistent results, confirming our architectural choice while validating AnchorDS's effectiveness across different conditioning mechanisms. IP-Adapter achieves superior performance in balancing the quality and 3D-consistency compared to other alternatives,
Our method demonstrates strong generalizability across different architectural configurations, indicating significant potential for integration with future advanced diffusion models and emerging conditioning mechanisms as they become available. 

\section{Discussion on Existing Methods under Our Formulation}\label{sec:discussion}
Our approach offers a new perspective on source conditioning in text-to-3D generation. It is insightful to compare \ourSDS\ with recent techniques that also aim to address SDS's oversmoothing artifacts.

\noindent\textbf{Comparison with SDS-Bridge.}
SDS-Bridge~\cite{mcallister2024sdsbridge} also recognizes the source distribution mismatch and proposes to use a negative prompt to describe the flaws of the current 3D model. In effect, SDS-Bridge replaces the unconditional score with a negatively-conditioned one (\eg, a prompt describing ``a bad, unfinished rendering''), then takes a difference similar to ours. However, this approach has fundamental limitations. A negative prompt is a \emph{static} descriptor and may not accurately characterize the 3D state as it evolves. If the chosen negative prompt diverges from actual errors in renders, guidance can become misaligned, sometimes pushing results further off-track. Indeed, SDS-Bridge is typically applied only after a period of normal SDS optimization, to ensure the 3D model reaches a ``rough'' state that the negative prompt can plausibly describe. In contrast, \ourSDS\ uses the rendered image $I^{(\tau)}$ itself as the descriptor of the current state, which is by definition precise and up-to-date. By conditioning on $I^{(\tau)}$ at every iteration, our source estimate adjusts automatically as the 3D asset changes—treating the 3D generation as dynamic distribution alignment rather than one-shot static correction. This dynamic anchoring eliminates the need for separate SDS pre-runs or hand-crafted prompts describing the source; the model's render provides all necessary information. Empirically, we found \ourSDS\ to be more robust than SDS-Bridge, which can fail when negative prompts are inadequate.

\noindent\textbf{Comparison with ProlificDreamer.} ProlificDreamer~\cite{wang_prolificdreamer_2023} takes a different approach: rather than reweighting a pretrained model's guidance, it trains a specialized diffusion branch via LoRA finetuning to better represent the current 3D scene. Their VSD guidance is defined as the difference $\hat{\boldsymbol\epsilon}_\phi(z_t; t, y_c)-\hat{\boldsymbol\epsilon}_{\phi}^{\text{(LoRA)}}(z_t; t, y, c)$, where $c$ represents view conditioning.
While VSD also yields difference-of-noise guidance, the philosophy diverges significantly from ours. VSD fine-tunes a LoRA model to directly approximate the evolving source distribution in latent space, treating the source as a particle distribution of 3D parameters (implicitly aggregating multi-view images). Due to the dynamic nature of the source distribution, VSD's latent space approximation inherently lags behind the actual distribution. Each fine-tuning step provides only slow incremental updates, resulting in persistent inaccuracies since the evolving distribution is never captured in time.

In contrast, \ourSDS\ does not aim to alter the diffusion model's internal latent distribution. Instead, it utilizes the diffusion model's existing capability to interpret and leverage image conditions. Fine-tuning an adapter in \ourSDS\ serves solely to familiarize the model with the conditional generation scenario rather than continuously updating internal distributions. Consequently, our fine-tuning is lightweight, converges quickly, and enables accurate, generalized estimation directly from the dynamically provided condition. This makes estimation inherently precise and immediately responsive to changes, avoiding the lag inherent in VSD's approach.

\noindent\textbf{Orthogonal Directions.} Several recent works address complementary aspects of SDS optimization. Methods like FSD~\cite{yan2024flowsd} and SSD~\cite{tang2023stable} focus on variance reduction in the noise estimator $\epsilon$. ISM~\cite{liang2024luciddreamer} addresses single-step score estimation inaccuracy through DDIM inversion for multi-step approximation. These approaches are orthogonal to our source distribution estimation focus and could potentially be combined with \ourSDS\ for further improvements.

\section{Comparison with Feed-forward and Hybrid Methods}

\begin{figure*}[tbh]
    \centering
    \includegraphics[width=0.9\linewidth]{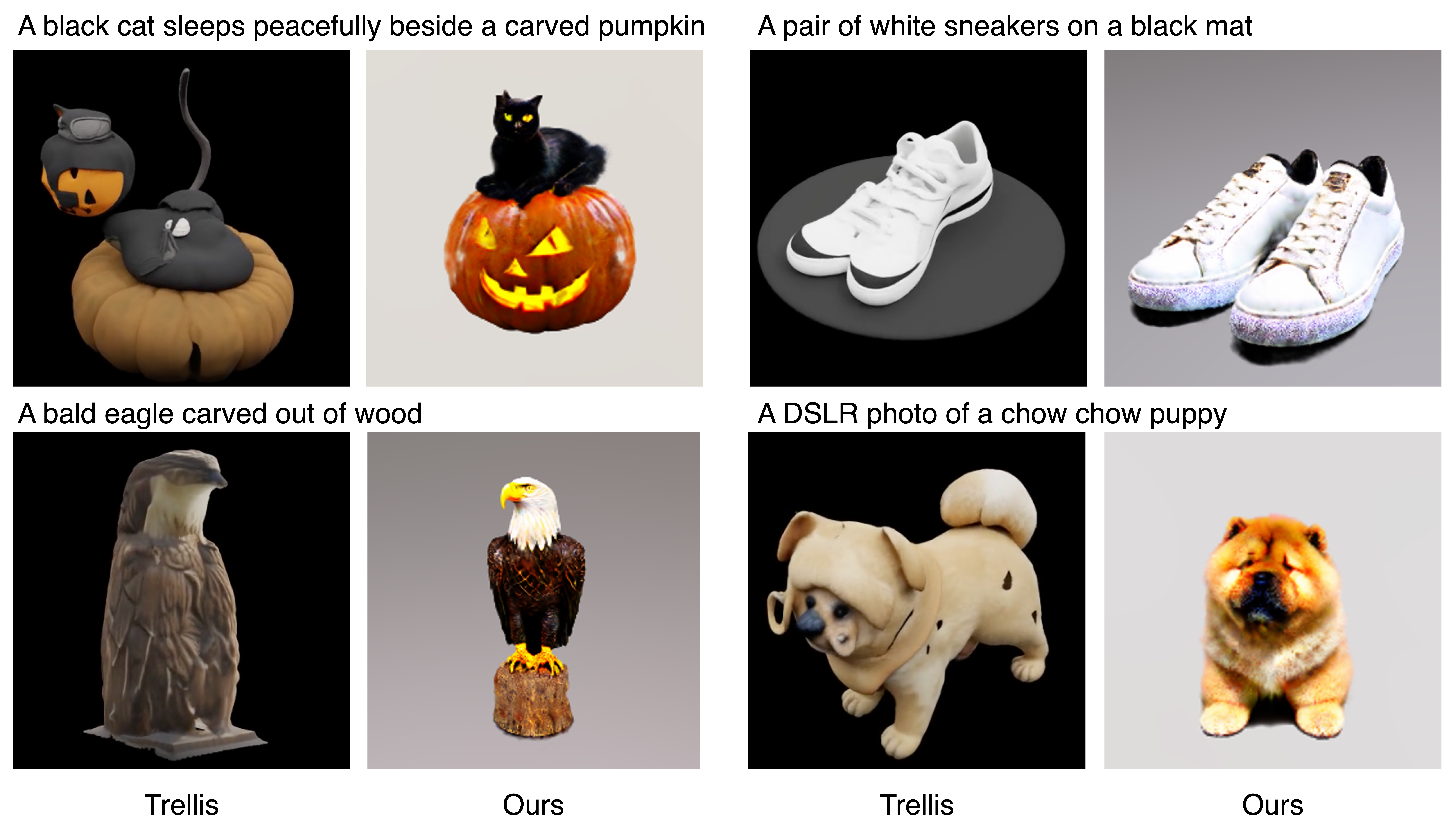}
    \caption{\small \textbf{Qualitative comparison with Trellis.} Our results demonstrate superior geometric and color fidelity, while Trellis suffers from out-of-distribution generalization issues due to its reliance on specific training datasets~\cite{he2023t3bench}, and exhibits poor performance under text prompt conditions~\cite{xiang2024trellis}.}
    \label{fig:appen-comp-trellis}
\end{figure*}

\noindent\textbf{Qualitative Comparison with Trellis.} 
We qualitatively compare our method with Trellis~\cite{xiang2024trellis}, a 3D latent-based text-to-3D model. As shown in Fig.~\ref{fig:appen-comp-trellis}, our results demonstrate superior geometric and color fidelity. In contrast, Trellis suffers from out-of-distribution generalization issues due to its reliance on 3D training datasets~\cite{he2023t3bench}, and exhibits poor performance under text prompt conditions~\cite{xiang2024trellis}.

\noindent\textbf{Quantitative Comparisons.}
Beyond qualitative comparisons, we report CLIP-based text-image alignment scores on our 50-prompt benchmark in Table~\ref{tab:ff_comparison}. We compare representative feed-forward text-to-3D methods against our \ourSDS\ initialized from Shap-E~\cite{jun2023shap_e}. \ourSDS\ achieves substantially higher CLIP similarity than Trellis~\cite{xiang2024trellis}, Shap-E, and Hunyuan3D-2.1, indicating stronger prompt fidelity despite operating as an SDS-based refinement. This supports our claim that dynamic source anchoring not only stabilizes SDS optimization but also surpasses existing feed-forward pipelines in semantic alignment.

Taken together, these quantitative results clarify the role of SDS-based optimization as a complementary paradigm to feed-forward pipelines. Feed-forward 3D generators are attractive for their fast inference, but are constrained by the coverage of their training corpora and often degrade on out-of-distribution or compositionally challenging prompts~\cite{he2023t3bench}. In contrast, SDS operates directly on arbitrary 3D parameterizations and can be deployed as a retraining-free refinement module that post-improves feed-forward outputs. %
By stabilizing SDS through our state-anchored guidance, \ourSDS\ prevents drift and oversmoothing, turning SDS from a fragile heuristic into a robust mechanism for high-fidelity, controllable 3D generation.

\begin{table}[t]
\centering
\scriptsize
\caption{\small \textbf{Quantitative comparison with feed-forward and hybrid methods.} CLIP-based image-text similarity on our 50-prompt set. \ourSDS\ consistently outperforms representative feed-forward and hybrid approaches, demonstrating that dynamic source anchoring yields superior semantic alignment while retaining the flexibility of SDS-style optimization.}
\begin{tabular}{lcc}
\toprule
Method & CLIP image-text sim $\uparrow$ \\
\midrule
Shap-E~\cite{jun2023shap_e} & 0.25 \\
Trellis~\cite{xiang2024trellis} & 0.22 \\
Hunyuan3D-2.1~\cite{hunyuan3d2025hunyuan3d}  & 0.29 \\
\ourSDS\ (ours) & \textbf{0.37} \\
\bottomrule
\end{tabular}
\label{tab:ff_comparison}
\end{table}

\section{More Qualitative Results}
\begin{figure*}[tbh]
    \centering
    \includegraphics[width=0.9\linewidth]{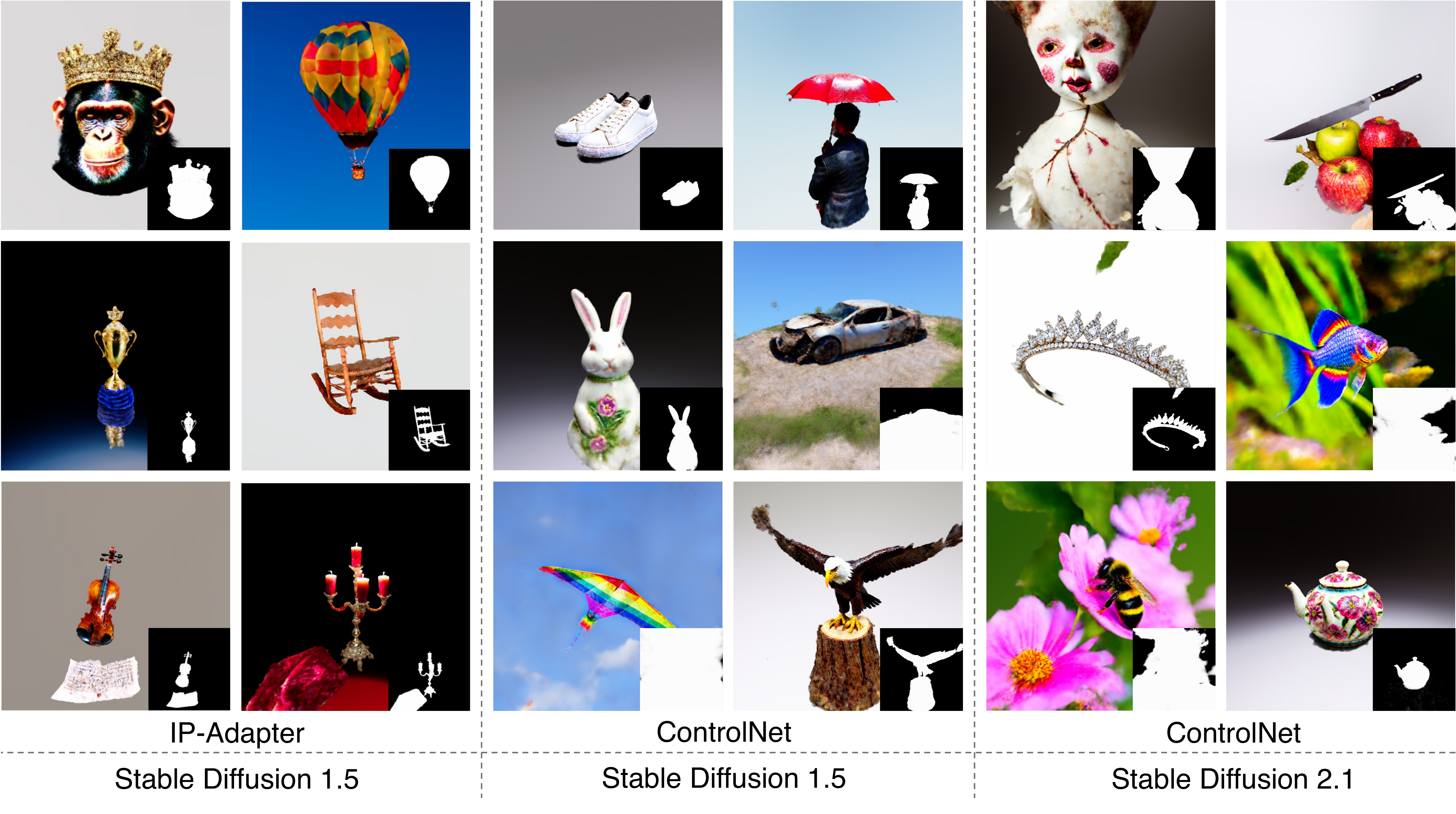}
    \caption{\small\textbf{More qualitative results on text-to-NeRF.} Prompts include: ``A chimpanzee dressed like Henry VIII king of England'', ``A hot air balloon in a clear sky'', ``A pair of white sneakers on a black mat'',  ``A man is holding an umbrella against rain'', ``A cracked porcelain doll's face'', ``Ripe apples cluster next to a gleaming knife'', ``The golden trophy shines brightly next to a ruffled blue ribbon'', ``A wooden rocking chair on a porch'', ``A small porcelain white rabbit figurine'', ``A completely destroyed car'', ``A sparkling diamond tiara'', ``A beautiful rainbow fish'',  ``A violin reclines on a chair next to a music sheet filled with notes'', ``A DSLR photo of a candelabra with many candles on a red velvet tablecloth'', ``A rainbow-colored kite soaring in the sky'', ``A bald eagle carved out of wood'', ``A bumblebee sitting on a pink flower'', ``A ceramic teapot with floral patterns''.}
    \label{fig:qual-nerf}
\end{figure*}

\begin{figure*}[tbh]
    \centering
    \includegraphics[width=0.9\linewidth]{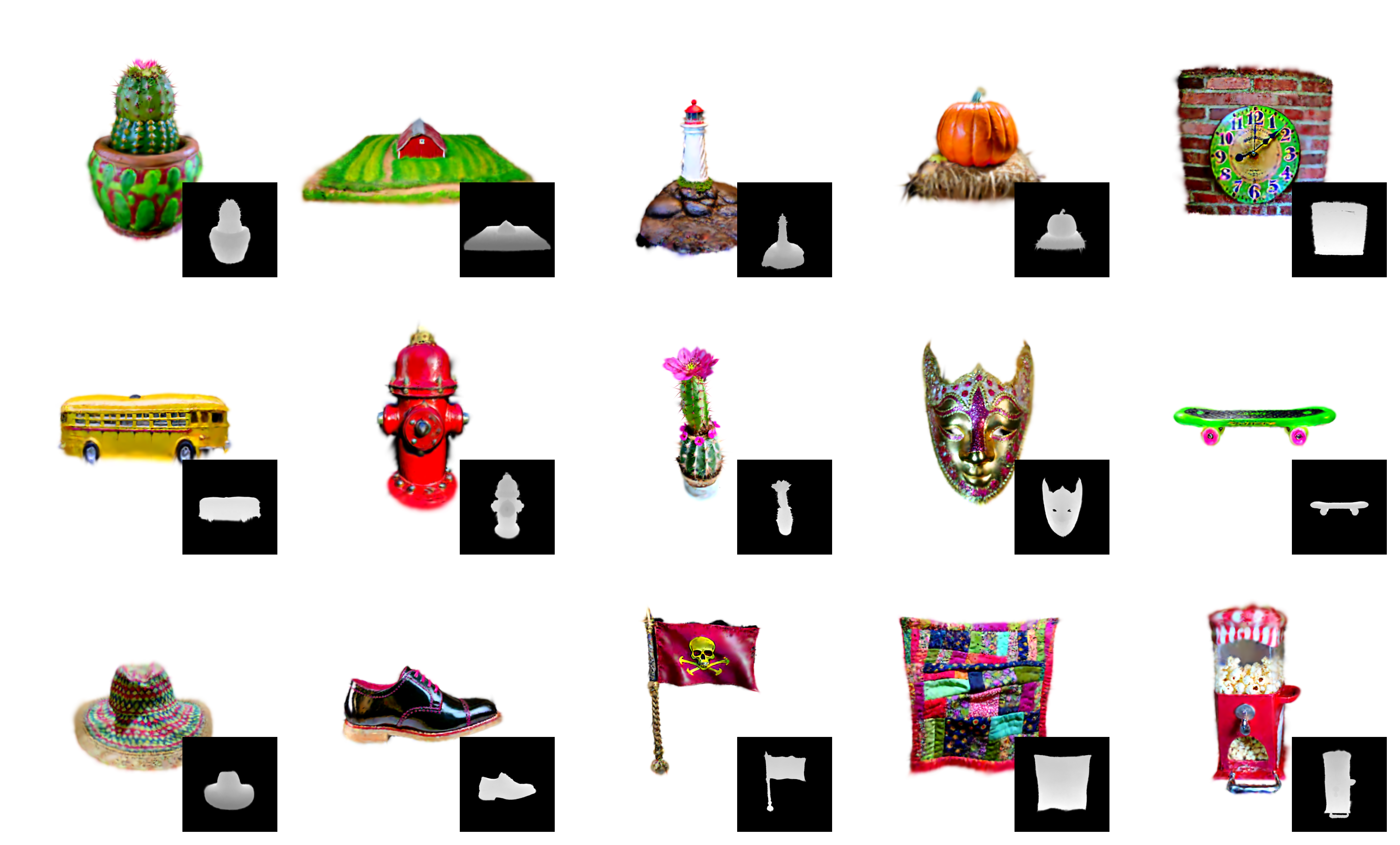}
    \caption{\small \textbf{More qualitative results on text-to-3DGS.} The figure showcases 3D object generation results using various prompts: ``A green cactus in a clay pot'', ``A red barn in a green field'', ``A lighthouse on a rocky shore'', ``A vibrant orange pumpkin sitting on a hay bale'', ``A vintage clock hanging on a brick wall'', ``A yellow school bus on a city street'', ``A bright red fire hydrant'', ``A cactus with pink flowers'', ``A gold glittery carnival mask'', ``A neon green skateboard with black wheels'', ``A well-worn straw sun hat'', ``A pair of shiny black patent leather shoes'', ``A pirate flag with skull and crossbones'', ``A vibrant, handmade patchwork quilt'', and ``Hot popcorn jump out from the red striped popcorn maker''. Each result demonstrates the method's ability to generate diverse 3D objects with varying complexity, materials, and semantic categories.}
    \label{fig:qual-3dgs}
\end{figure*}

Additional qualitative results for NeRF (Fig.~\ref{fig:qual-nerf}) and 3DGS (Fig.~\ref{fig:qual-3dgs}) showcase consistent performance across diverse semantic categories and complexity levels.


\section{Details of User Study}\label{appen:user-study}
\begin{figure*}[!htbp]
    \centering
    \includegraphics[width=\linewidth]{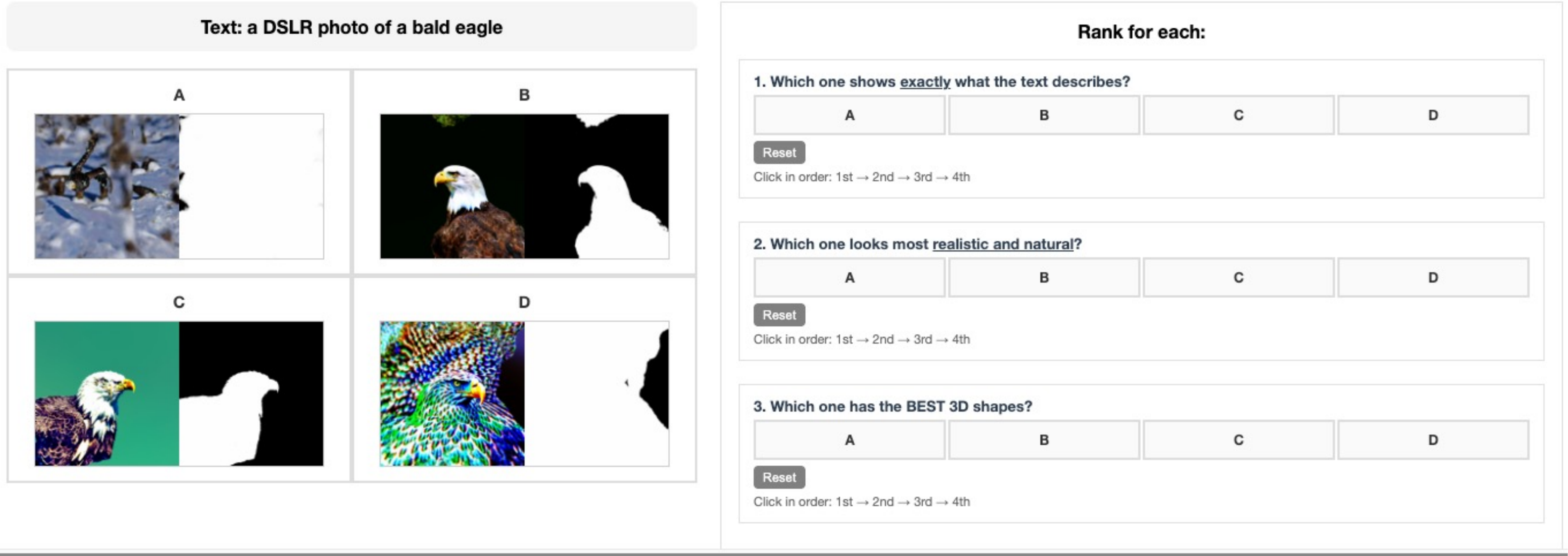}
    \caption{Snapshot of the user study interface. Participants were shown the prompt and rendered images generated by different methods. They were asked to rank the methods following their preferences based on overall quality, 3D consistency, and consistency with the prompt.}
    \label{fig:appen-userstudy}
\end{figure*}

We conducted a human preference study to evaluate the effectiveness of our proposed AnchorDS compared to existing methods using all $50$ complex prompts. A total of 912 unique participants were recruited through Amazon Mechanical Turk, resulting in 1000 effective comparison samples in total. %
The interface is shown in Fig.~\ref{fig:appen-userstudy}. For each comparison, participants were provided with the text prompt and the randomly ordered generationed results of various methods. They were then asked to indicate their preference rankings based on three evaluation criteria:
\begin{itemize}
    \item 3D Consistency: The output that maintains the best 3D consistency;
    \item Test Consistency: The output that better maintains consistency with the input prompt;
    \item Visual Quality: The output with the best overall visual quality.
\end{itemize}
The methods are grouped as SD 2.1-based and SD 1.5-based, and evaluated separately. For the SD 2.1-based methods, the users are simply asked to indicate their most preferred result, since there are just two methods to compare.

\section{Limitations and Future Work} %
Our method shares several limitations common to SDS-based approaches. In particular, the quality of the generated 3D content is highly dependent on the guidance from 2D prior models; if the underlying 2D model fails to generate meaningful representations, the corresponding 3D outputs are also compromised. This limitation could be alleviated by leveraging more powerful 2D backbones, such as SD3 or Flux.1. Additionally, we observe that convergence stability is sensitive to the choice of 3D representation, suggesting room for improvement in representation learning and rendering fidelity.

On the score distillation side, our work focuses primarily on addressing the first type of error—source estimation bias—as discussed in SDS-Bridge. We leave the integration of methods to mitigate the second class of error—trajectory estimation mismatch—as future work. Promising directions include the use of inversion-based techniques or improved sampling strategies to more accurately track the evolving distribution across denoising steps.

Finally, our key contribution lies in reinterpreting 3D generation as an evolving editing process. This perspective opens up new avenues for future research, such as unifying the formulation of 3D generation and 3D editing within a single framework, and extending our approach to more controllable or user-driven editing tasks.

\end{document}